\documentclass[letterpaper, 10 pt, conference]{ieeeconf}  

\IEEEoverridecommandlockouts                              

\usepackage{cite}
\usepackage{amsmath,amssymb,amsfonts}
\usepackage{mathtools}
\usepackage{algorithmic}
\usepackage{graphicx}
\usepackage{textcomp}
\usepackage{float}
\usepackage[hyphens]{url}
\usepackage{hyperref}
\usepackage[hyphenbreaks]{breakurl}
\usepackage[capitalise]{cleveref}
\usepackage[table,xcdraw]{xcolor}
\usepackage{physics}
\usepackage{tikz,pgf} 

\usepackage{cuted} 

\usepackage{mathdots}
\usepackage{yhmath}
\usepackage{cancel}
\usepackage{color}
\usepackage{siunitx}
\usepackage{array}
\usepackage{multirow}

\usepackage{gensymb}
\usepackage{tabularx}
\usepackage{extarrows}

\usepackage[caption=false]{subfig}

\usepackage{booktabs}

\usepackage[utf8]{inputenc}
\usepackage{pgfplots}
\usepgfplotslibrary{groupplots,dateplot}
\usetikzlibrary{fadings}
\usetikzlibrary{patterns,shapes.arrows}
\usetikzlibrary{shadows.blur}
\usetikzlibrary{shapes}
\pgfplotsset{compat=newest}

\usetikzlibrary{external}
\usetikzlibrary{calc}

\usepackage{bbm}  
\renewcommand{\mathbb}{\mathbbm}

\def\permille{\ensuremath{{}^\text{o}\mkern-5mu/\mkern-3mu_\text{oo}}}

\def\useTiksExtPDF{1} 
\usepackage{xstring} 

\newcommand{\includetikz}[3]{%

    \def\fn{#2}
    \StrSubstitute{\fn}{ }{_}[\fn]

    \StrSubstitute{\fn}{\%}{_}[\fn]

    \ifx#31 
        {\tikzexternaldisable
            \input{#1/#2.tikz}%
        }
    \else
        \ifx\useTiksExtPDF\undefined 
            \tikzsetnextfilename{output_\fn}%
            \input{#1/#2.tikz}%
        \else
            \if\useTiksExtPDF1  
                \includegraphics[]{tikz/output_\fn}
            \else  
                \tikzsetnextfilename{output_\fn}%
                \input{#1/#2.tikz}%
            \fi
        \fi
    \fi

}

\title{\LARGE \bf
A Probabilistic Programming Idiom for Active Knowledge Search
}
\author{Malte R. Damgaard \and Rasmus Pedersen \and Thomas Bak
\thanks{All authors are with Department of Electronic Systems, Automation and Control, Aalborg University,     Denmark,{\tt\small (e-mail:~\{mrd, rpe, tba\}@es.aau.dk)}.}%
}

\begin{document}

\maketitle
\thispagestyle{empty}
\pagestyle{empty}

\begin{abstract}
In this paper, we derive and implement a probabilistic programming idiom for the problem of acquiring new knowledge about an environment. The idiom is implemented utilizing a modern probabilistic programming language. We demonstrate the utility of this idiom by implementing an algorithm for the specific problem of active mapping and robot exploration. Finally, we evaluate the functionality of the implementation through an extensive simulation study utilizing the HouseExpo dataset.
\end{abstract}

\section{Introduction}\label{sec:introduction}

Making decisions under uncertainty to obtain new knowledge about an environment is a recurring problem within robotics.
To efficiently solve this problem, the robot needs to continuously learn about its environment while keeping track of the uncertainty about current knowledge. The decision-making is further complicated if an extrinsic reward signal cannot guide the robot and if predefined constraints should be satisfied.

The most well-known and studied problem of this type within robotics is probably active mapping and robot exploration \cite{s21072445}. Most solutions to active mapping and robot exploration heavily exploit the structure of the stored knowledge, i.e., the map, to derive efficient algorithms. E.g. for grid map representations, it is common to apply frontier exploration \cite{topiwala2018frontier,7276723,9429053}. These methods exploit the property, that it is possible to identify frontiers between the knowledge represented by grid cells in a grid map, that the robot is currently certain about, and the knowledge for which it is uncertain. Actions are chosen to guide the robots towards these frontiers, by which the map is explored. While such approaches exploiting problem-specific properties can result in efficient solutions, they do not easily generalize to other types of knowledge. 
E.g. because such exploration frontiers cannot easily be defined for other types of knowledge.

Other solutions to active mapping and robot exploration take a deep-learning approach, to learn an efficient policy for acquiring new knowledge. E.g., in \cite{li2019houseexpo} they feed the current knowledge, again in the form of a grid map, into an artificial neural network and let the output of the network control the actions of the robot. They then train the network with a reward equal to the newly discovered area at each time-step, by which they obtain a policy for exploration. While such an approach can be very efficient at specific tasks, end-to-end learning often limits the generalizability of the solution due to a lack of structural transferability. In many cases, the artificial neural network would have to be re-trained to work for other problems requiring other inputs and outputs.


Opposite to the problem-specific approaches already mentioned, the goal of cognitive architectures is to create computational entities with general problem-solving capabilities, that should function across a multitude of tasks. In recent years a community consensus about the overall structure and components of cognitive architectures has begun to emerge, called the Standard Model of the Mind \cite{Laird_Lebiere_Rosenbloom_2017}. Especially, the realization of the need for an efficient combination of symbolic and statistical processing is a massive change compared to early research in cognitive architectures. 
In \cite{damgaard2021idiomatic} we presented a generalized framework for developing such cognitive architectures for robotics applications. This was done in an effort to standardize work and promote better cooperation. One of the main ideas of the framework is to develop and identify general and reusable fragments of probabilistic programs, i.e., probabilistic programming idioms, for which inference could be done efficiently utilizing variational inference methods.


Inspired by some of the main concepts of the Standard Model of the Mind, the goal of the presented efforts is to develop such a general and reusable probabilistic programming idiom for the problem of making decisions under uncertainty to obtain new knowledge about an environment. The main contributions of this paper are:
\begin{enumerate}
    \item Derivation and implementation of the said probabilistic programming idiom, 
    \item and validation of the said idiom used in an active mapping and robot exploration context through simulations on a large dataset.
\end{enumerate}
We choose to validate the idiom based on the active mapping and robot exploration problem because it is a well-studied problem with a relatively simple problem formulation for which results are easily interpretable via visual inspection of the robot's trajectory. Still, the problem is sufficiently hard due to the non-convex constraints implied by objects in the environment.

\Cref{sec:Preliminaries} presents preliminaries necessary to understand the content of the following sections. In \cref{sec:decision_model} the derivation of the probabilistic programming idiom is presented.
In \cref{sec:Autonomous_Robot_Exploration} the application of the idiom for the active mapping and robot exploration problem is presented, together with the results of an extensive simulation study. Finally, in \cref{sec:discussion} we conclude upon the presented work, and hint to future lines of research.

\section{Preliminaries}\label{sec:Preliminaries}
Within this paper $Z$ is used to denote latent variables, $X$ is used to denote observed variables, and $C$ is used to denote a collection of both types of variables. We use a superscript in curly brackets to indicate the index of a variable. Specially, for time indexes, we indicate the set of indexes of future variables as $\left\{t\right\}^+ = \left\{t+1,...,t+\overline{T}\right\}$. Similarly, we indicate the set of indexes of past variables as $\left\{t\right\}^- = \left\{t-\underline{T},...,t\right\}$. We develop our model primarily for approximate inference with stochastic variational inference. In general, variational inference refer to methods that approximates one conditional distribution, $p( z|x=\overline{x})$ with another unconditional distribution, $q(z)$, through an optimization problem of the form
\begin{align*}
\underset{q( z) \in Q}{\text{min}} D[ p( z|x=\overline{x}) ||q( z)] & 
\end{align*}
where $Q$ is the family of distributions from which $q$ can be picked, and $D$ is a divergence measure quantifying the difference between $p$ and $q$. In stochastic variational inference, $q$ is assumed to be parameterised by a set of parameters $\phi$, and stochastic gradient ascent is used to solve a tractable dual-problem \cite{JMLR:v14:hoffman13a}. To solve this dual-problem, we do not need to know the conditional distribution, $p( z|x=\overline{x})$, but only need to specify the unconditional model, $p( z, x=\overline{x})$, making it a lot easier to work with. However, to use stochastic variational inference we need to ensure that our unconditional model, $p( z, x=\overline{x})$, preserves the differentiability of the dual-problem. Within this paper, we will make use of divergence measures from the family of f-divergences, defined by
\begin{align*}
D_{f}[ p( z) ||q( z)] & =\int _{z} q( z) f\left(\frac{p( z)}{q( z)}\right)\\
 & =E_{q( z)}\left[ f\left(\frac{p( z)}{q( z)}\right)\right]
\end{align*}
where $f$ is an arbitrary convex function \cite{1633769}. Based on f-divergence we can define the f-information measure as
\begin{align*}
I_{f}[ z,y] & =D_{f}[ p( z) p( y) ||p( z,y)]\\
 & =E_{p( y)}[ D_{f}[ p( z) ||p( z|y)]].
\end{align*}
The commonly used KL-divergence, $D_{\text{KL}}$, and mutual information is defined by $f( u) =-log( u)$ such that
\begin{align*}
D_{\text{KL}}[ p( z) ||q( z)] & =E_{q( z)}[ log( q( z)) -log( p( z))]
\end{align*}
Similarly, the inverse-KL-divergence, $D_{\overline{KL}}$, and Lautum information, $I_{L}$, is defined by $\displaystyle f( u) =u\cdot log( u)$, from which we can obtain the conditional Lautum information measure
\begin{align}
I_{L}[z,y|x] & =E_{p( y|x)}[ D_{\overline{KL}}[ p( z|x) ||p( z|y,x)]] \label{eq:lautum_information}\\
 & =E_{p( y|x)}\left[\begin{matrix*}[l]
log( E_{p( z|x)}[ p( y|z,x)])\\
\quad -E_{p( z|x)}[ log( p( y|z,x))]
\end{matrix*}\right]
\end{align}
For more information about these measures and their properties, we refer the reader to \cite{1633769,minka2005divergence}. Within this paper, we will also be using the following approximate ”probabilistic logic”
\begin{align*}
& p( z\in \overline{z} \lor y\in \overline{y})\overset{\text{def}}{=}\\
& \quad\quad\quad p( z\in \overline{z}) +p( y\in \overline{y}) -p( z\in \overline{z}) \cdot p( y\in \overline{y})\\
& p( z\in \overline{z} \land y\in \overline{y})\overset{\text{def}}{=} p( z\in \overline{z}) \cdot p( y\in \overline{y})\\
& p\left(\bigwedge _{i=1}^{I} z^{\{i\}} \in \overline{z}^{\{i\}}\right)\overset{\text{def}}{=}\prod _{i=1}^{I} p\left( z^{\{i\}} \in \overline{z}^{\{i\}}\right)
\end{align*}
where we have used $\lor$ and $\land$ to denote the approximate \textit{or} and the \textit{and} operation, respectively. These approximate ”probabilistic logic” rules simply constitute a probabilistic union and intersection with an implied independence assumption, respectively.

\section{Decision model}\label{sec:decision_model}

\begin{figure}[!b]
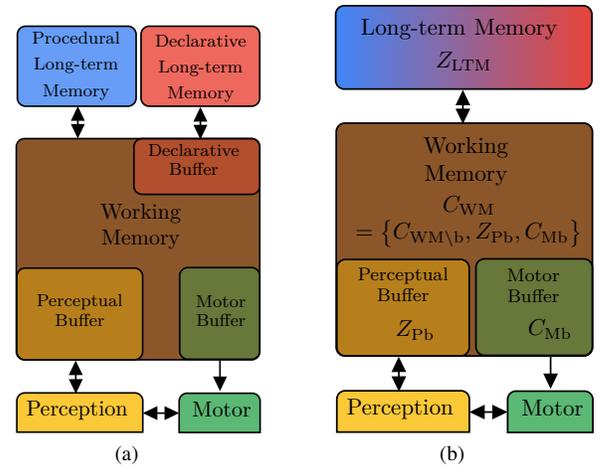

    \centering
    \subfloat[\label{fig:memory_structure_standard_model}]{
        \resizebox{0.46\linewidth}{!}{
            \includetikz{tikz/source_files}{memory_structure_standard_model}{0}
         }
    }
    \subfloat[\label{fig:memory_structure_in_this_paper}]{
        \resizebox{0.485\linewidth}{!}{
            \includetikz{tikz/source_files}{memory_structure_in_this_paper}{0}
         }
    }
    \caption{(a) The conceptual memory structure of the Standard Model \cite{Laird_Lebiere_Rosenbloom_2017}. The red, blue, brown, green, and yellow colourse are related to declarative long-term memory, procedural long-term memory, working memory, perception component, and motor component, respectively. Here we have used rectangles with rounded corners to symbolise pure memory, and sharp corners to indicate a relation to external signals. (b) The conceptual memory structure used within this paper is without the distinction between procedural and declarative long-term memory, and without the \textit{declarative buffer}. The figure also indicates the symbols used for each type of memory.}
    \label{fig:memory_structures}
\end{figure}


According to \cite{Laird_Lebiere_Rosenbloom_2017} it is commonly agreed that the memory structure of mind like architectures at a top-level conceptually can be divided into \textit{working memory} and \textit{long-term memory} each of which constitutes relations over symbols supplemented by quantitative metadata to provide a hybrid symbolic-subsymbolic representation. Besides the two main types of memory, it is also agreed that there exists an architectural component denoted \textit{perception} for converting external signals into appropriate memory representations. Similarly, there exists an architectural component denoted \textit{motor} for translating internal memory representations into external signals. The relations between each of the aforementioned are illustrated in \cref{fig:memory_structure_standard_model}. \textit{Long-term memory} is responsible for the storage of information over extended periods. The \textit{working memory} includes temporary information necessary for behavior production and problem-solving, such as information about goals, but also contains different buffers for temporarily storing information from the \textit{perception} component, the \textit{motor} component, and some types of long-term memory. As such \textit{working memory} acts as a linkage between the other components. It is customary to sub-divide long-term memory further into specialized types of memories. However, since we in this paper are focusing on decision making to acquire new knowledge, that is updating all types of long-term memory, we will not make such distinctions, as illustrated in \cref{fig:memory_structure_in_this_paper}. Neither will we make a distinction between the declarative buffer and general long-term memory, and jointly refer to them as long-term memory. Furthermore, to keep our presentation relatively concise, we will not consider the relation between working memory, and the \textit{perception} and \textit{motor} components. Instead we will assume that appropriate \textit{perception} and \textit{motor} components are present.

To accommodate the need for a hybrid symbolic-subsymbolic representation as suggested by the standard model, we will derive a probabilistic model of decision making. Based on the division of memory and the symbol definitions indicated in \cref{fig:memory_structure_in_this_paper} we make the following definition of the joint probability distribution
\begin{align} 
    &p\left( C_{\text{WM} \backslash \text{b}} ,C_\text{Mb} ,Z_\text{Pb},Z_\text{LTM}\right) \nonumber\\ 
    &= p\left( Z_{\text{WM} \backslash \text{b}}^{\left\{t\right\}^-}, C_{\text{WM} \backslash \text{b}}^{\left\{t\right\}^+}, X_\text{Mb}^{\left\{t-1\right\}^-}, Z_\text{Mb}^{\left\{t-1\right\}^+}, Z_\text{Pb}^{\left\{t\right\}^-} ,Z_\text{Pb}^{\left\{t\right\}^+}, Z_\text{LTM}\right) \nonumber\\ 
    & \overset{\text{def}}{=} \underbrace{p\left( C_{\text{WM} \backslash \text{b}}^{\left\{t\right\}^+} ,Z_\text{Mb}^{\left\{t-1\right\}^+} ,Z_\text{Pb}^{\left\{t\right\}^+} |\breve{Z}_{\text{WM} \backslash \text{b}}^{\left\{t\right\}^-} ,\breve{Z}_\text{LTM}\right)}_\text{Planning/Decision Making} \label{eq:ourCognition}\\
    & \qquad\qquad\qquad \cdot \underbrace{p\left( Z_{\text{WM} \backslash \text{b}}^{\left\{t\right\}^-}, X_\text{Mb}^{\left\{t-1\right\}^-} ,Z_\text{Pb}^{\left\{t\right\}^-}, Z_\text{LTM} \right)}_\text{Learning/Reasoning} \nonumber
\end{align}
where we have used sub-script "$\text{WM} \backslash \text{b}$" to denote the set of variables representing the working memory except of the set of variables representing the two buffers, i.e., $C_\text{Mb}$ and $Z_\text{Pb}$. In \cref{eq:ourCognition} we have assumed that the distribution of future variables, $C_{\text{WM}\backslash \text{b}}^{\left\{t\right\}^+}$, $Z_\text{Mb}^{\left\{t-1\right\}^+}$, and $Z_\text{Pb}^{\left\{t\right\}^+}$ are conditional independent of previous information in the perceptual buffer, $Z_\text{Pb}^{\left\{t\right\}^-}$, and motor buffer, $X_\text{Mb}^{\left\{t-1\right\}^-}$, given the previous variables in the rest of the working memory, $Z_{\text{WM}\backslash \text{b}}^{\left\{t\right\}^-}$, and the long-term memory, $Z_\text{LTM}$. The last fraction of \cref{eq:ourCognition} deals with inference over variables internal to an agent based on past experience in the form of the variables of the perceptual buffer, $Z_\text{Pb}^{\left\{t\right\}^-}$, and the motor buffer, $Z_\text{Mb}^{\left\{t-1\right\}^-}$, related to the past. As such this fraction corresponds to reasoning and \textit{learning}. Similarly, the first factor of \cref{eq:ourCognition} only deals with future variables based on what have already been learned from past experiences. Since it is assumed that the working memory includes information necessary for behavior production this fraction is responsible for decision making and \textit{planning} guided by preferences contained in the working memory. By the nature of the problem, the probabilistic causation between \textit{learning} and \textit{planning} should only be one way, from \textit{learning} to \textit{planning}. In other words, we can consider inference over the variables in the \textit{learning} part in isolation, and when performing inference in the \textit{planning} part we should keep the \textit{learning} distribution fixed. 
To emphasize this, we have used breves over the variables ${Z}_{\text{WM} \backslash \text{b}}^{\left\{t\right\}^-}$ and ${Z}_\text{LTM}$ in the first fraction of \cref{eq:ourCognition}. 
The proposed model effectively divides the cognitive tasks of an agent into \textit{learning} and \textit{planning}. Assuming that we have access to the \textit{learning} distribution, this allows us to focus the rest of the paper on the \textit{planning} part.

For the purpose of decision making, and to make our model resemble the classical Markov decision process, we introduce the following variables as a part of the working memory. State variables, $Z_{\text{s}}$, representing the state of the agent itself and the environment. Decision variables, $C_{\text{D}}$, explicitly represent preferences such as goals and constraints. That is, $Z_{\text{s}}^{\left\{t\right\}^-}  \in C^{\left\{t\right\}^-}_{\text{WM} \backslash \text{b}}$ and $\{Z_{\text{s}}^{\left\{t\right\}^+} ,C_{\text{D}}^{\left\{t\right\}^+}\} \in C^{\left\{t\right\}^+}_{\text{WM} \backslash \text{b}}$. Furthermore, adopting the Markov property between state variables also used in the Markov decision process we define the planning distribution from \cref{eq:ourCognition} as
\begin{align}
    & p\left( C_{\text{WM} \backslash \text{b}}^{\{t\}^{+}} ,Z_{\text{Mb}}^{\{t-1\}^{+}} ,Z_{\text{Pb}}^{\{t\}^{+}} |\breve{Z}_{\text{WM} \backslash \text{b}}^{\{t\}^{-}} ,\breve{Z}_{\text{LTM}}\right)\nonumber\\
    & \ \overset{def}{=}\prod _{\tau =t+2}^{t+\overline{T}}\left[\begin{array}{ c }
        \begin{matrix*}[l]
        p\left( C_{\text{D}}^{\{\tau \}} |Z_{\text{s}}^{\{\tau \}} ,\breve{Z}_{\text{WM} \backslash \text{b}}^{\{t\}^{-}} ,\breve{Z}_{\text{LTM}}\right)\\
        \quad \cdot p\left( Z_{\text{s}}^{\{\tau \}} |Z_{\text{s}}^{\{\tau -1\}} ,Z_{\text{Mb}}^{\{\tau -1\}}\right) p\left( Z_{\text{Mb}}^{\{\tau -1\}}\right)
        \end{matrix*}
    \end{array}\right]\nonumber\\
    & \qquad \cdot p\left( C_{\text{D}}^{\{t+1\}} |Z_{\text{s}}^{\{t+1\}} ,\breve{Z}_{\text{WM} \backslash \text{b}}^{\{t\}^{-}} ,\breve{Z}_{\text{LTM}}\right)\label{eq:ourPlanning}\\
    & \qquad \cdot p\left( Z_{\text{s}}^{\{t+1\}} |\breve{Z}_{\text{s}}^{\{t\}} ,Z_{\text{Mb}}^{\{t\}}\right) p\left( Z_{\text{Mb}}^{\{t\}}\right) \nonumber
\end{align}


The causality structure of \cref{eq:ourPlanning} goes as follows. 
The current possible content of the motor buffer, $Z_{\text{MB}}^{\{\tau \}}$, together with the belief over the state at that time instance, $Z_{\text{s}}^{\{\tau \}}$, determines the belief over the next possible states, $Z_{\text{s}}^{\{\tau+1 \}}$. 
The next possible states, $Z_{\text{s}}^{\{\tau+1 \}}$, together with the variables in the long-term memory, $Z_{\text{LTM}}$, and all variables related to the past in the working memory except the buffers, $Z_{\text{WM} \backslash \text{b}}^{\{t\}^{-}}$, potentially contributes to the current belief over the decision variables, $Z_{\text{D}}^{\{\tau \}}$. Except for the decision variables and the explicit inclusion of the long-term memory variables, most parts of \cref{eq:ourPlanning} resembles elements known from other decision models such as the Partially observable Markov decision process. As stated earlier, the decision variables are meant to guide the decision process, and as such might be problem-dependent. 

\begin{figure}[!ht]
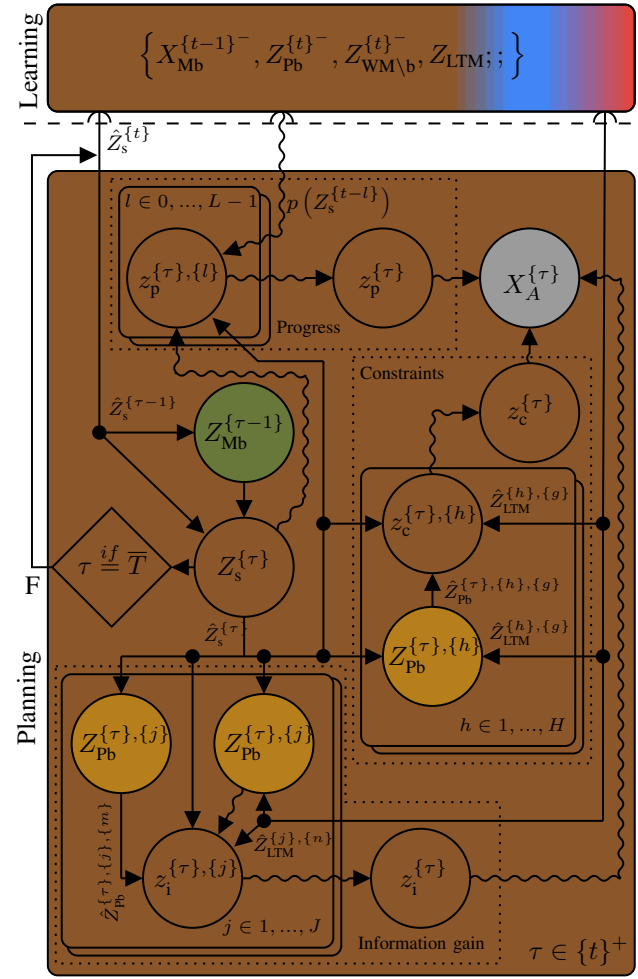

\centering
\includetikz{tikz/source_files}{generative_flow_graph_3}{1}
\caption{Illustration of the generative flow of the proposed exploration idiom. Rectangles with rounded corners represent a collection of variables. Two stacked rectangles with rounded corners represent conditional independent collections of variables. Circles indicate a distribution over the variable inside the circle. They thereby constitute potential samples sites in the probabilistic program. Solid arrows indicate samples passed around in the probabilistic program, sampled from the distribution represented by the circle at the origin of the arrow. Wavy arrows indicate an evaluation at a specific point of the parent distribution represented by the circle at the origin of the arrow. Gray-colored circles indicate "observed" variables, while other colors are used to indicate a relation to the different types of memories. The rectangles with dotted borders indicate the variables associated with the three different decision variables.}\label{fig:generative_flow_graph_3}
\end{figure}

For the purpose of making decisions in order to obtain new knowledge, and inspired by \cite{10.1007/978-3-319-21365-1_15} we chose to include and combine the following general purpose decision variables: progress, $z_{\text{p}}$, information gain, $z_{\text{i}}$, constraint, $z_{\text{c}}$, and attention, $x_{\text{A}}$. From these we define $C_{\text{D}}^{\{\tau \}} =\{x_{\text{A}}^{\{\tau \}}, z_{\text{p}}^{\{\tau \}}, z^{\{\tau \}}_{\text{i}}, z_{\text{c}}^{\{\tau \}}\}$. The meaning of these variables are described in the following sections. For reference the structure of the combined model is indicated in \cref{fig:generative_flow_graph_3}.

\subsection{Progress}
The progress variable is meant to quantify how different a given state, $Z^{\{\tau \}}_{\text{s}}$, is from the past states,  $Z^{\{\tau \}^{-}}_{\text{s}}$. To quantify the progress while taking uncertainty into account we can make use of the divergence measures described in \cref{sec:Preliminaries}. However, calculating such divergence measures inside a probabilistic program amounts to a form of nested inference which potentially can cause problems. E.g. when we want to use stochastic variational inference as the main inference algorithm we have to make sure that we can calculate the gradient of the nested inference performed. Here we choose to use the following one-point estimate of the KL-divergence as a measure of progress
\begin{align}
 & D_{\text{KL}}\left[ p\left( Z_{\text{s}}^{\{t-l\}}\right) ||p\left( Z_{\text{s}}^{\{\tau \}} |Z_{\text{s}}^{\{\tau -1\}} ,Z_{\text{Mb}}^{\{\tau -1\}}\right)\right]\nonumber\\
 & =E_{\hat{Z}_{\text{s}}^{\{\tau \}}}\left[ log\left(\frac{p\left( Z_{\text{s}}^{\{\tau \}} =\hat{Z}_{\text{s}}^{\{\tau \}} |Z_{\text{s}}^{\{\tau -1\}} ,Z_{\text{Mb}}^{\{\tau -1\}}\right)}{p\left( Z_{\text{s}}^{\{t-l\}} =\hat{Z}_{\text{s}}^{\{\tau \}}\right)}\right)\right]\nonumber\\
 & \approx \frac{1}{I}\sum _{i=1}^{I}\left(\begin{matrix}
log\left( p\left( Z_{\text{s}}^{\{t-l\}} =\hat{Z}_{\text{s}}^{\{\tau \} ,\{i\}}\right)\right)\\
-log\left( p\left( Z_{\text{s}}^{\{\tau \}} =\hat{Z}_{\text{s}}^{\{\tau \} ,\{i\}} |Z_{\text{s}}^{\{\tau -1\}} ,Z_{\text{Mb}}^{\{\tau -1\}}\right)\right)
\end{matrix}\right)\nonumber\\
 & \approx ReLu\left(\begin{matrix}
log\left( p\left( Z_{\text{s}}^{\{t-l\}} =\hat{Z}_{\text{s}}^{\{\tau \}}\right)\right)\\
-log\left( p\left( Z_{\text{s}}^{\{\tau \}} =\hat{Z}_{\text{s}}^{\{\tau \}} |Z_{\text{s}}^{\{\tau -1\}} ,Z_{\text{Mb}}^{\{\tau -1\}}\right)\right)
\end{matrix}\right)\nonumber\\
 & \overset{\text{def}}{=} P^{\{t-l\}}\left(\hat{Z}_{\text{s}}^{\{\tau \}}\right) \label{eq:progress_approximation}
\end{align}
Where $p\left( Z_{\text{s}}^{\{t-l\}}\right)$ is a marginal of the learning distribution in \cref{eq:ourCognition}, $\displaystyle \hat{Z}_{\text{s}}^{\{\tau \}} \sim p\left( Z_{\text{s}}^{\{\tau \}} |Z_{\text{s}}^{\{\tau -1\}} ,Z_{\text{Mb}}^{\{\tau -1\}}\right)$ and $\displaystyle \hat{Z}_{\text{s}}^{\{\tau \} ,\{j\}} \sim p\left( Z_{\text{s}}^{\{\tau \}} |Z_{\text{s}}^{\{\tau -1\}} ,Z_{\text{Mb}}^{\{\tau -1\}}\right)$, and we have used the ReLu function in our approximation since $\displaystyle log\left(\frac{p_{1}}{p_{2}}\right) \ngeq 0$ in general but $\displaystyle D_{\text{KL}}[ p_{1} ||p_{2}] \geq 0$. The gradient of the log-probability function can be calculated for many commonly used distributions and probabilistic programs composed of these, and thereby also for this approximation. From this approximation we define the distribution over the progress variable for a given state, $Z^{\{\tau \}}_{\text{s}}$, relative to a single of the past states,  $Z^{\{t-l \}}_{\text{s}}$, as
\begin{align}
 & p\left( z_{\text{p}}^{\{\tau \} ,\{l\}} |Z_{\text{s}}^{\{\tau \}} =\hat{Z}_{\text{s}}^{\{\tau \}}\right) =\label{eq:progress_single}\\
\  & \ \ Bernoulli\left( \lambda _{p}^{\{l\}} \cdot \left[ 1-e^{-\sigma _{p} \cdot P^{\{t-l\}}\left(\hat{Z}_{\text{s}}^{\{\tau \}}\right)}\right]\right)\nonumber
\end{align}
where
\begin{align*}
\lambda _{p}^{\{l\}} & =1-\frac{( 1-\lambda _{\text{p,min}})( L-1-l)}{L-1} \quad ;\begin{matrix}
L >1,\\
L\leqslant \underline{T} ,
\end{matrix}
\end{align*}
is a decay variable used to put more emphasis on the oldest states considered, L is the number of old states considered, and $\displaystyle \sigma _{p}$ is simply a scaling parameter. Here we have used a trick commonly utilised in probabilistic Reinforcement Learning, and Control \cite{DBLP:journals/corr/abs-1805-00909}, where a given reward is converted to a pseudo probability by exponentiation of that reward. Since, $P^{\{t-l\}}\left(\hat{Z}_{\text{s}}^{\{\tau \}}\right)\geq 0$ it follows that $e^{-\sigma _{p} \cdot P^{\{t-l\}}\left(\hat{Z}_{\text{s}}^{\{\tau \}}\right)}\in[0,1]$ and thus it can be used as a pseudo probability. \cref{eq:progress_single} thus state that
a state, $Z_{\text{s}}^{\{\tau \}}$, yielding a higher approximated divergence, $P^{\{t-l\}}\left(\hat{Z}_{\text{s}}^{\{\tau \}}\right)$, has an exponentially higher probability of yielding progress. From \cref{eq:progress_single} we define the total progress as the combined progress relative to all of the last $L\leq t-\underline{T}$ past states
\begin{align}
 & p\left( z_{\text{p}}^{\{\tau \}} |Z_{\text{s}}^{\{\tau \}} =\hat{Z}_{\text{s}}^{\{\tau \}}\right) =\label{eq:progress}\\
\  & \ \ Bernoulli\left( p\left(\left.\bigwedge _{l=0}^{L-1} z_{\text{p}}^{\{\tau \} ,\{l\}} =1 \right|Z_{\text{s}}^{\{\tau \}} =\hat{Z}_{\text{s}}^{\{\tau \}}\right)\right) \nonumber
\end{align}
The approximation in \cref{eq:progress_approximation} might seem very coarse; however, when used as nested inference inside a stochastic variational inference algorithm, it is evaluated multiple times during inference of the main problem. The effect is thus effectively similar to a mean approximation using many samples.

\subsection{Information Gain}
As the name implies, the information gain variable, $z_{i}$, is meant to quantify the amount of information that can potentially be gained from being in a specific state, $Z^{\{\tau \}}_{\text{s}}$, perceiving the environment and thereby obtain new information through the perceptual buffer. The perceptual buffer might contain information from multiple independent perceptual modalities, which we will denote as $\displaystyle Z_{\text{Pb}}^{\{\tau \} ,\{j\}}$. Each of these perceptual modalities might only relate to a specific part of the long-term memory which we will denote $\displaystyle Z_{\text{LTM}}^{\{j\}}$. To quantify the expected amount of information obtained by being in a specific state, $Z^{\{\tau \}}_s$, we use the Lautum information in \cref{eq:lautum_information}. Based on the Lautum information we represent the pseudo probability that the perceptual modality, $\displaystyle Z_{\text{Pb}}^{\{\tau \} ,\{j\}}$, will yield new knowledge as the distribution
\begin{align}
 & p\left( z_{\text{i}}^{\{\tau \} ,\{j\}} |Z_{\text{s}}^{\{\tau \}} =\hat{Z}_{\text{s}}^{\{\tau \}}\right) = \label{eq:information_j}\\
\  & \ \ Bernoulli\left( 1-e^{-\sigma _{I} \cdot I_{\text{L}}\left[ Z_{\text{LTM}}^{\{j\}} ,Z_{\text{Pb}}^{\{\tau \} ,\{j\}} |Z_{\text{s}}^{\{\tau \}} =\hat{Z}_{\text{s}}^{\{\tau \}}\right]}\right) \nonumber
\end{align}
where $\displaystyle \sigma _{I}$ is a scaling parameter. Maximising the information obtained by each of the perceptual modalities might require wildly different changes to the state, $Z^{\{\tau \}}_{\text{s}}$. Therefore, we focus the attention on the modality providing the most information and define
\begin{align*}
 & p\left( z_{\text{i}}^{\{\tau \}} |Z_{\text{s}}^{\{\tau \}} =\hat{Z}_{\text{s}}^{\{\tau \}}\right) =\\
\  & \ \ Bernoulli\left(\underset{j\in [ 1,J]}{max} \ p\left( z_{\text{i}}^{\{\tau \} ,\{j\}} =1|Z_{\text{s}}^{\{\tau \}} =\hat{Z}_{\text{s}}^{\{\tau \}}\right)\right).
\end{align*}
Calculating the Lautum information inside a probabilistic program also amounts to nested inference. To make the calculation of Lautum information compatible with the use of stochastic variational inference for the main problem, we again make use of the following sample mean estimate
\begin{align*}
& I_{\text{L}}[ x,y|z=\hat{z}] =E_{\hat{y}}\left[\begin{matrix*}[l]
log( E_{\hat{x}}[ p( y=\hat{y} |x=\hat{x} ,z=\hat{z})])\\
\quad -E_{\hat{x}}[ log( p( y=\hat{y} |x=\hat{x} ,z=\hat{z}))]
\end{matrix*}\right] \\
 & \approx \frac{1}{M}\sum\limits_{m=1}^{M}\left[\begin{matrix*}[l]
log\left(\frac{1}{N}\sum\limits_{n=1}^{N} p\left( y=\hat{y}^{\{m\}} |x=\hat{x} ,z=\hat{z}\right)\right)\\
-\frac{1}{N}\sum\limits_{n=1}^{N} log\left( p\left( y=\hat{y}^{\{m\}} |x=\ \hat{x}^{\{n\}} ,z=\hat{z}\right)\right)
\end{matrix*}\right]\\
 & =\frac{1}{M}\sum\limits_{m=1}^{M}\left[\begin{matrix*}[l]
log\left(\sum\limits_{n=1}^{N} e^{log\left( p\left( y=\hat{y}^{\{m\}} |x=\ \hat{x}^{\{n\}} ,z=\hat{z}\right)\right)}\right)\\
-log( N)\\
-\frac{1}{N}\sum\limits_{n=1}^{N} log\left( p\left( y=\hat{y}^{\{m\}} |x=\ \hat{x}^{\{n\}} ,z=\hat{z}\right)\right)
\end{matrix*}\right]
\end{align*}
where $\displaystyle \hat{x} \sim p( x|z=\hat{z})$ and $\displaystyle \hat{y} \sim p( y|z=\hat{z})$, \ $\displaystyle \hat{y}^{\{m\}} \sim p( y|z=\hat{z})$ and $\displaystyle \ \hat{x}^{\{n\}} \sim p( x|z=\hat{z})$. To use the approximation in \cref{eq:information_j}, we only need to be able to evaluate 
\begin{equation*}
log\left( p\left( Z_{\text{Pb}}^{\{\tau \} ,\{j\}} =\hat{Z}_{\text{Pb}}^{\{\tau \} ,\{j\} ,\{m\}} \left|\begin{matrix}
Z_{\text{s}}^{\{\tau \}} =\hat{Z}_{\text{s}}^{\{\tau \}} ,\\
Z_{\text{LTM}}^{\{j\}} =\hat{Z}_{\text{LTM}}^{\{j\} ,\{n\}}
\end{matrix}\right. \right)\right)
\end{equation*}
with 
\begin{align*}
\hat{Z}_{\text{LTM}}^{\{j\} ,\{n\}} & \sim p\left( Z_{\text{LTM}}^{\{j\}}\right)\\
\hat{Z}_{\text{Pb}}^{\{\tau \} ,\{j\} ,\{m\}} & \sim p\left( Z_{\text{Pb}}^{\{\tau \} ,\{j\}} |Z_{\text{s}}^{\{\tau \}} =\hat{Z}_{\text{s}}^{\{\tau \}}\right)
\end{align*}
where we have assumed that $\displaystyle p\left( Z_{\text{LTM}}^{\{j\}} |Z_{\text{s}}^{\{\tau \}}\right) =p\left( Z_{\text{LTM}}^{\{j\}}\right)$.

\subsection{Constraints}
The constraint variable, $z_{c}^{\{\tau \}}$, is meant to quantify states, $Z^{\{\tau \}}_\text{s}$, that should be avoided taking perceived information, $Z_\text{PB}^{\{\tau \}}$, and knowledge stored in long-term memory, $Z_\text{LTM}$, into account. Often such constraints can be defined by a set, $\displaystyle A^{\{\tau \} ,\{h\}}$, that the state, $\displaystyle Z_{\text{s}}^{\{\tau \}}$, should be within. 
As this set might depend on knowledge stored in the long-term memory, $\displaystyle Z_{\text{LTM}}$, and the expected content of the perceptual buffer, $\displaystyle Z_{\text{Pb}}^{\{\tau \}}$, we assume a set definition of the form
\begin{align*}
 & A^{\{\tau \} ,\{h\}} =\\
 & \quad \left\{Z_{\text{s}}^{\{\tau \}} ,Z_{\text{Pb}}^{\{\tau \}} ,Z_{\text{LTM}} \ |\ \mathbb{1}_{A}^{\{\tau \} ,\{h\}}\left( Z_{\text{s}}^{\{\tau \}} ,Z_{\text{Pb}}^{\{\tau \}} ,Z_{\text{LTM}}\right)\right\}
\end{align*}
where $\displaystyle \mathbb{1}_{A}^{\{\tau \} ,\{h\}}$ is the indicator function of the set $\displaystyle A^{\{\tau \} ,\{h\}}$. Given that $\displaystyle Z_{\text{s}}^{\{\tau \}} =\hat{Z}_{\text{s}}$ the probability that the constraint defined by the set $\displaystyle A^{\{\tau \} ,\{h\}}$ is satiesfied can then be expressed as
\begin{align}
 & P\left( Z_{\text{s}}^{\{\tau \}} =\hat{Z}_{\text{s}}^{\{\tau \}} ,Z_{\text{Pb}}^{\{\tau \}} ,Z_{\text{LTM}} \in A^{\{\tau \} ,\{h\}}\right) \label{eq:constraint_probability}\\
 & \quad \ \ =E_{\hat{Z}_{\text{Pb}}^{\{\tau \}} ,\hat{Z}_{\text{LTM}}}\left[\mathbb{1}_{A}^{\{\tau \} ,\{h\}}\left(\hat{Z}_{\text{s}}^{\{\tau \}} ,\hat{Z}_{\text{Pb}}^{\{\tau \}} ,\hat{Z}_{\text{LTM}}\right)\right] \nonumber
\end{align}
where $\displaystyle \hat{Z}_{\text{Pb}}^{\{\tau \}} \sim p\left( Z_{\text{Pb}}^{\{\tau \}} |\hat{Z}_{\text{s}}^{\{\tau \}} =\hat{Z}_{\text{s}}^{\{\tau \}} ,\hat{Z}_{\text{LTM}} =\hat{Z}_{\text{LTM}}\right)$ and $\displaystyle \hat{Z}_{\text{LTM}} \sim p( Z_{\text{LTM}})$. Based on this we define distribution over the constraint variable for the $h$'th constraint at time $\tau$ as
\begin{align*}
 & p\left( z_{\text{c}}^{\{\tau \} ,\{h\}} |Z_{\text{s}}^{\{\tau \}} =\hat{Z}_{\text{s}}^{\{\tau \}}\right) =\\
\  & \quad Bernoulli\left( P\left( Z_{\text{s}}^{\{\tau \}} =\hat{Z}_{\text{s}}^{\{\tau \}} ,Z_{\text{Pb}}^{\{\tau \}} ,Z_{\text{LTM}} \in A^{\{\tau \} ,\{h\}}\right)\right)
\end{align*}
and distribution over the combined constraint variable at time $\tau$ as
\begin{align*}
 & p\left( z_{\text{c}}^{\{\tau \}} |Z_{\text{s}}^{\{\tau \}} =\hat{Z}_{\text{s}}^{\{\tau \}}\right) =\\
 & \ \ \ Bernoulli\left( p\left(\left.\bigwedge _{h=1}^{H} z_{\text{c}}^{\{\tau \} ,\{h\}} =1 \right|Z_{\text{s}}^{\{\tau \}} =\hat{Z}_{\text{s}}^{\{\tau \}}\right)\right)
\end{align*}
Where $\displaystyle H$ is the number of constraints. Calculating the probability in \cref{eq:constraint_probability} also amounts to nested inference, but the discontinuity of the indicator function for the set definition, $\displaystyle \mathbb{1}_{A}^{\{\tau \} ,\{h\}}$, also present a problem for calculating the gradients needed for stochastic variational inference. To overcome this, we assume that the indicator function can be specified as
\begin{align*}
 & \mathbb{1}_{A}^{\{\tau \} ,\{h\}}\left( Z_{\text{s}}^{\{\tau \}} ,Z_{\text{Pb}}^{\{\tau \}} ,Z_{\text{LTM}}\right)\\
 & \quad\quad=\begin{cases}
1 & d^{\{\tau \} ,\{h\}}\left( Z_{\text{s}}^{\{\tau \}} ,Z_{\text{Pb}}^{\{\tau \}} ,Z_{\text{LTM}}\right)  >0\\
0 & else
\end{cases}
\end{align*}
and make the approximation
\begin{align*}
\mathbb{1}_{A}^{\{\tau \} ,\{h\}}\left( Z_{\text{s}}^{\{\tau \}}\right) & \approx \tilde{\mathbb{1}}_{A}^{\{\tau \} ,\{h\}}\left( d^{\{\tau \} ,\{h\}}\left( Z_{\text{s}}^{\{\tau \}} ,Z_{\text{Pb}}^{\{\tau \}} ,Z_{\text{LTM}}\right)\right)
\end{align*}
where $\displaystyle \tilde{\mathbb{1}}_{A}^{\{\tau \} ,\{h\}}( x) \in [ 0,1]$ is a smooth monotonically increasing function symmetric around $\displaystyle \tilde{\mathbb{1}}_{A}^{\{\tau \} ,\{h\}}( 0) =0.5$, e.g., a scaled sigmoid function. From this we make use the sample mean approximation to obtain the following approximation to the probability in \cref{eq:constraint_probability}
\begin{align*}
 & P\left( Z_{\text{s}}^{\{\tau \}} =\hat{Z}_{\text{s}}^{\{\tau \}} ,Z_{\text{Pb}}^{\{\tau \}} ,Z_{\text{LTM}} \in A^{\{\tau \} ,\{h\}}\right)\\
 & \quad \ \ \approx \frac{1}{G}\sum _{g=1}^{G}\mathbb{1}_{A}^{\{\tau \} ,\{h\}}\left(\hat{Z}_{\text{s}}^{\{\tau \}} ,\hat{Z}_{\text{Pb}}^{\{\tau \} ,\{g\}} ,\hat{Z}_{\text{LTM}}^{\{g\}}\right)\\
 & \quad \ \ \approx \frac{1}{G}\sum _{g=1}^{G}\tilde{\mathbb{1}}_{A}^{\{\tau \} ,\{h\}}\left( d^{\{\tau \} ,\{h\}}\left(\hat{Z}_{\text{s}}^{\{\tau \}} ,\hat{Z}_{\text{Pb}}^{\{\tau \} ,\{g\}} ,\hat{Z}_{\text{LTM}}^{\{g\}}\right)\right)
\end{align*}
where $\displaystyle \hat{Z}_{\text{Pb}}^{\{\tau \} ,\{g\}} \sim p\left( Z_{\text{Pb}}^{\{\tau \}} |\hat{Z}_{\text{s}}^{\{\tau \}} ,\hat{Z}_{\text{LTM}}\right)$ and $\displaystyle \hat{Z}_{\text{LTM}}^{\{g\}} \sim p( Z_{\text{LTM}})$.

\subsection{Attention}
Finally, the attention variable is meant to summarise the other decision variables and symbolises which states the agent should focus its attention on. Based on the approximate "probabilistic logic" presented in \cref{sec:Preliminaries} we define.
\begin{align}
    & p\left( x_{A}^{\tau } |Z_{\text{s}}^{\{\tau \}} =\hat{Z}_{\text{s}}^{\{\tau \}}\right) =\label{eq:attention}\\
    &\quad Bernoulli\left(
        p\left(\left.\begin{matrix*}[l]
            \left[\begin{matrix*}[l]
            z_{\text{p}}^{\{\tau \}} =1\\
            \quad\lor z_{\text{i}}^{\{\tau \}} =1
            \end{matrix*}\right]\\
            \quad\qquad\land z_{\text{c}}^{\{\tau \}} =1
        \end{matrix*} \right| Z_{\text{s}}^{\{\tau \}} =\hat{Z}_{\text{s}}^{\{\tau \}}\right)\right) \nonumber
\end{align}
Basically, \cref{eq:attention} states that an agent should focus its attention on states that either yields progress, \textit{or} yield new knowledge, \textit{and} also satisfies the given constraints. 

\subsection{Variational distribution}
As stated in \cref{sec:Preliminaries}, a parameterized unconditional variational distribution, $q\left(Z_{\text{WM}}^{\{t\}^{+}}\right)$, needs to be specified to utilize stochastic variational inference for approximate inference. Most of the factors in \cref{eq:ourPlanning} are assumed to be known and thus fixed. Therefore, only the distribution over the variables in the motor buffer can be considered a free distribution, and thus we define
\begin{align*}
q\left( Z_{\text{WM}}^{\{t\}^{+}}\right) & =q\left( Z_{\text{WM} \backslash \text{b}}^{\{t\}^{+}} ,Z_{\text{Mb}}^{\{t-1\}^{+}}\right)\\
& \overset{def}{=} p\left( Z_{\text{s}}^{\{t+1\}} |\breve{Z}_{\text{s}}^{\{t\}} ,Z_{\text{Mb}}^{\{t\}}\right) q_{\phi _{\text{Mb}}^{\{\tau -1\}}}\left( Z_{\text{Mb}}^{\{t\}} |Z_{\text{s}}^{\{t\}}\right).\\
& \quad \quad \cdot \prod _{\tau =t+2}^{t+\overline{T}}\left[\begin{array}{ c }
\begin{matrix*}[l]
p\left( Z_{\text{s}}^{\{\tau \}} |Z_{\text{s}}^{\{\tau -1\}} ,Z_{\text{Mb}}^{\{\tau -1\}}\right)\\
\ \cdot q_{\phi _{\text{Mb}}^{\{\tau -1\}}}\left( Z_{\text{Mb}}^{\{\tau -1\}} |Z_{\text{s}}^{\{\tau -1\}}\right)
\end{matrix*}
\end{array}\right]
\end{align*}

where $\phi _{\text{Mb}}^{\{\tau\}}$ are the parameters that need to be found by stochastic variational inference.

\begin{figure*}[!ht]
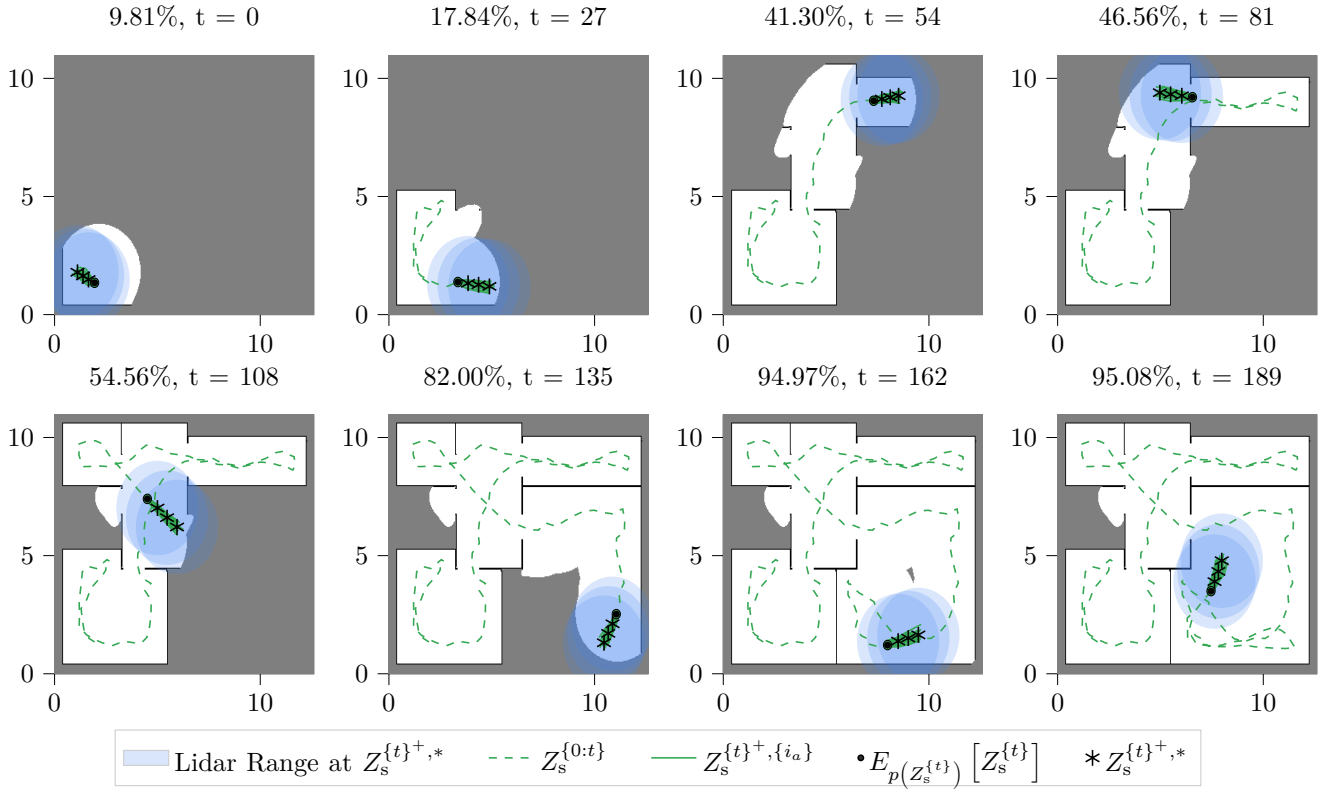

\centering
\includetikz{tikz/source_files/graphs/good_map_example}{good_map_example}{0}
\caption{The progress of a simulation for the map with ID "7fb9c9203cb8c4404f4af1781f1c6999" after each of the timesteps $t=\left\{0,27,54,81,108,135,162,189\right\}$. For each timestep the previous positions, $Z_{\text{s}}^{\{0:t\}}$, is shown by a green dashed line, the mean of the current position, $Z_{\text{s}}^{\{t\}}$, is shown by a black dot, samples of future positions, $Z_{\text{s}}^{\{\tau \}^{+} ,\{i_{a}\}}$, is shown by solid green lines, and the mean of these samples corresponding to the optimal future positions based on $Z_{\text{a}}^{\{t\}^{+},*}$ are shown by black asterisks. The simulation where terminated after $t=189$ since the exploration percentage where above $95\%$.}\label{fig:good_map_example}
\end{figure*}

\subsection{Summery}
So far, general functionality that could potentially be utilised for multiple problems has been described and thus could be considered an idiom. This idiom is implemented as an abstract class utilising the probabilistic programming language Pyro \cite{bingham2019pyro} developed on top of PyTorch and python. The class contains the following abstract methods that need to be implemented
\begin{table}[H]
    \centering\def\arraystretch{2}
    \begin{tabular}{m{0.4\linewidth}|m{0.4\linewidth}}
    $\displaystyle q_{\phi _{\text{Mb}}^{\{\tau -1\}}}\left( Z_{\text{Mb}}^{\{\tau -1\}} |Z_{\text{s}}^{\{\tau -1\}}\right)$ & $\displaystyle p\left( Z_{\text{Pb}}^{\{\tau \},\{j\}} |Z_{\text{s}}^{\{\tau \}} ,Z_{\text{LTM}}^{\{j\}}\right)$ \\
    $\displaystyle p\left( Z_{\text{Mb}}^{\{\tau -1\}} |Z_{\text{s}}^{\{\tau -1\}}\right)$ & $\displaystyle d^{\{\tau \},\{h\}}\left( Z_{\text{s}}^{\{\tau \}} ,Z_{\text{Pb}}^{\{\tau \}} ,Z_{\text{LTM}}\right)$ \\
    $\displaystyle p\left( Z_{\text{s}}^{\{\tau \}} |Z_{\text{s}}^{\{\tau -1\}} ,Z_{\text{Mb}}^{\{\tau -1\}}\right)$ & $\displaystyle \tilde{\mathbb{1}}_{A}^{\{\tau \},\{h\}}( d)$ \\
    $\displaystyle p\left( Z_{\text{Pb}}^{\{\tau \},\{j\}} |Z_{\text{s}}^{\{\tau \}}\right)$ & $p\left( Z_{\text{LTM}}^{\{j\}}\right)$ \\
    \end{tabular}
\end{table}
The abstract methods representing probability functions need to be implemented as compatible probabilistic programs utilising Pyro. Besides the abstract methods users also need to provide $p\left( Z_{\text{s}}^{\{t\}}\right)$ as a probabilistic program. Besides the above necessary methods, the class also specifies two additional methods that can be used to control the sub-sampling of the long-term memory and perceptual buffer for use in the calculation of information gain and constraint violations. With these methods implemented users can call the class method "\textit{makePlan(...)}" which via stochastic variational inference finds an approximate optimal set of parameters, $q_{\phi _{\text{Mb}}^{\{\tau \},*}}$ to the variational inference problem 
\begin{align*}
\underset{\phi _{\text{WM}}^{\{t\}^{+}}}{\text{min}} D\left[\begin{matrix}
p\left( Z_{\text{WM}}^{\{t\}^{+}}\left| \begin{matrix}
\breve{Z}_{\text{s}}^{\{t\}} ,\breve{Z}_{\text{LTM}} ,\\
x_{A}^{\{t\}^{+}} =1
\end{matrix}\right. \right)\\
\cdot p\left( Z_{\text{WM}}^{\{t\}^{-}} ,Z_{\text{LTM}}\right)
\end{matrix}\right| \left| \begin{matrix}
q_{\phi _{\text{WM}}^{\{t\}^{+}}}\left( Z_{\text{WM}}^{\{t\}^{+}}\left| \begin{matrix}
\breve{Z}_{\text{s}}^{\{t\}} ,\\
\breve{Z}_{\text{LTM}}
\end{matrix}\right. \right)\\
\cdot p\left( Z_{\text{WM}}^{\{t\}^{-}} ,Z_{\text{LTM}}\right)
\end{matrix}\right]. 
\end{align*}
where the user can specify the divergence measure and optimiser used. The optimal set of parameters is used to draw samples of the future motor buffer
\begin{align*}
    Z_{\text{Mb}}^{\{\tau \}} \sim q_{\phi _{\text{Mb}}^{\{\tau \},*}}\left( Z_{\text{Mb}}^{\{\tau \}} |Z_{\text{s}}^{\{\tau \}}\right) \quad \quad ;\tau \in \{t,...,\overline{T}-1\}.
\end{align*}
These samples constitute potential future optimal actions needed to optimise information gain or progress while satisfying constraints. Finally, the "\textit{makePlan(...)}" method either returns these samples or a sample mean hereof. The code is available trough \cite{Damgaard_probMind_2022}.


\section{Autonomous Robot Exploration}\label{sec:Autonomous_Robot_Exploration}
To exemplify the utility of the proposed idiom, we have used it to implement an algorithm for autonomous robot exploration. The code for this can be found through \cite{Damgaard_probMind_2022}. The goal of the implementation is for a robot to explore an environment represented by a grid map autonomously, consider the problem at a high level, and define the state to be the current position in the XY-plane, $Z_{\text{s}}^{\{\tau \}} =\left[ z_{\text{x}}^{\{\tau \}} ,z_{\text{y}}^{\{\tau \}}\right]^{T}$, and use the simple transition model as
\begin{align*}
p\left( Z_{\text{s}}^{\{\tau +1\}} |Z_{\text{s}}^{\{\tau \}} ,Z_{\text{Mb}}^{\{\tau \}}\right) & =N\left( Z_{\text{s}}^{\{\tau \}} +A\left( Z_{\text{Mb}}^{\{\tau \}}\right) ,\sigma _{\text{a}}\right)
\end{align*}
where $\displaystyle Z_{\text{Mb}}^{\{\tau \}}$ is the relative position scaled to be in the interval $\displaystyle \left[ 0,1\right]$, $\displaystyle A( ...)$ is a linear scaling of the relative position to be in the range $\displaystyle \left[\underline{\Delta Z_{\text{a}}} ,\overline{\Delta Z_{\text{a}}}\right]$, and $\displaystyle \sigma _{\text{a}}$ is the covariance of the error allowed in the movement. Since the robot should have no prior preference of its movement we define
\begin{align}
p\left( Z_{\text{Mb}}^{\{\tau \}} |Z_{\text{s}}^{\{\tau \}}\right) & =U\left(\left[\begin{matrix}
0\\
0
\end{matrix}\right] ,\left[\begin{matrix}
1\\
1
\end{matrix}\right]\right) \label{eq:p_z_Mb}.
\end{align}
$\displaystyle q_{\phi _{\text{Mb}}^{\{\tau \}}}\left( Z_{\text{Mb}}^{\{\tau \}} |Z_{\text{s}}^{\{\tau \}}\right)$ should have the same support as \cref{eq:p_z_Mb}, but should also be flexible enough to represent preferences in the relative position. Thus, we define
\begin{align*}
q_{\phi _{\text{Mb}}^{\{\tau \}}}\left( Z_{\text{Mb}}^{\{\tau \}} |Z_{\text{s}}^{\{\tau \}}\right) & =Beta\left( \alpha _{\text{a}}^{\{\tau \}} ,\beta _{\text{a}}^{\{\tau \}}\right)
\end{align*}
since the beta distribution subsumes the uniform distribution, but also can represent a single mode. Thus, we have $\displaystyle \phi _{\text{Mb}}^{\{\tau \}} =\left\{\alpha _{\text{a}}^{\{\tau \}} ,\beta _{\text{a}}^{\{\tau \}}\right\}$. We consider the grid map to be the long-term memory. That is, $Z_{\text{LTM}} =\left\{z_{m}^{\{1\}} ,...,z_{m}^{\{I_{m}\}}\right\}$ where $z_{m}^{\{i_{m}\}}$ is each of the cells in the grid map, and make the common assumption that
\begin{align*}
p( Z_{\text{LTM}}) & =\prod _{i_{m} =1}^{I} p\left( z_{m}^{\{i_{m}\}}\right)
\end{align*}
where
\begin{align*}
p\left( z_{m}^{\{i_{m}\}}\right) & =Bernoulli\left( P_{m}^{\{i_{m}\}}\right)
\end{align*}
and $P_{m}^{\{i_{m}\}}$ is the probability of the $i_{m}$'th grid cell being occupied. We assume that the environment is perceived through a lidar with $\displaystyle 360\degree $ field of view and evenly spaced lidar beams with $\displaystyle 1\degree $ spacing, and define 
\begin{align*}
p\left( Z_{\text{Pb}}^{\{\tau \}} |Z_{\text{s}}^{\{\tau \}} ,Z_{\text{LTM}}\right) & =\prod _{1=i_{r}}^{360} p\left( z_{\text{r,d}}^{\{\tau \},\{i_{r} \}} |Z_{\text{s}}^{\{\tau \}} ,Z_{\text{LTM}}^{\{\tau \},\{i_{r} \}}\right)
\end{align*}
where $\displaystyle z_{\text{r,d}}^{\{\tau \},\{i_{r} \}}$ is the distance measured by the $\displaystyle i_{r}$'th laser beam at time $\displaystyle \tau $ given the current position and grid map, and 
\begin{align*}
Z_{\text{LTM}}^{\{\tau \},\{i_{r} \}} & =\left\{z_{m}^{\{i_{m}\}} \in Z_{\text{LTM}} |\text{ray} \ i_{r}\text{ intersects cell } i_{m}\right\}
\end{align*}
is the cell in the grid map that the $\displaystyle i_{r}$'th laser beam intersects. We obtain the set, $\displaystyle Z_{\text{LTM}}^{\{\tau \},\{i_{r} \}}$, through ray-tracing. The distribution $p\left( z_{\text{r,d}}^{\{\tau \},\{i_{r} \}} |Z_{\text{s}}^{\{\tau \}} ,Z_{\text{LTM}}^{\{\tau \},\{i_{r} \}}\right)$ is implemented according to the laser beam model in \cite{probRobotics}. Without taking the map into consideration the robot have no prior knowledge on the distance measured by the lidar, and thus we define
\begin{align*}
p\left( z_{\text{r,d}}^{\{\tau \},\{i_{r} \}} |Z_{\text{s}}^{\{\tau \}}\right) & =U\left( 0,\overline{z_{\text{r,d}}}\right)
\end{align*}
where $\displaystyle \overline{z_{\text{r,d}}}$ is the max range of the lidar beams. We furthermore want the robot to keep a minimum distance, $\displaystyle d_{min}$, to occupied cells in the map and thus define the constraints via the logistic function
\begin{align*}
 & \tilde{\mathbb{1}}_{A}^{\{\tau \},\{h\}}\left( d^{\{\tau \},\{h\}}\left( Z_{\text{s}}^{\{\tau \}} ,Z_{\text{Pb}}^{\{\tau \}} ,Z_{\text{LTM}}\right)\right)\\
 & \qquad\qquad\qquad\qquad\qquad=\frac{1}{1+e^{-\sigma _{c}\cdot\left( z_{\text{r,d}}^{\{\tau \},\{h\}} -d_{min}\right)}}
\end{align*}
where $\displaystyle \sigma _{c}$ determines the steepness of the logistic function. With the above definitions, we have $\displaystyle J=H=360$. Calculating the information gain and constraint violation based on all 360 lidar beams is computationally intractable in the current implementation. Therefore, for each timestep, $\displaystyle \tau $, we sub-sample the number of lidar beams taken into account by randomly picking $\displaystyle \tilde{J} \ll 360$ and $\displaystyle \tilde{H} \ll 360$ lidar beams for calculating the information gain and constraint violation, respectively. In our implementation, we have furthermore chosen to use Pyro's build-in "ClippedAdam" optimizer with the standard $D_{\text{KL}}$ divergence measure. Finally, the next action that the robot should take, $Z_{\text{a}}^{\{t\},*}$, is calculated as the sample mean of optimal actions
\begin{align}
Z_{\text{a}}^{\{t\},*} & =\frac{1}{I_{a}}\sum _{i_{a} =1}^{I_{a}} A\left( Z_{\text{Mb}}^{\{t\},\{i_{a}\}}\right) \label{eq:optimal_action}
\end{align}
where $\displaystyle Z_{\text{Mb}}^{\{\tau \},\{i_{a}\}} \sim q_{\phi _{\text{Mb}}^{\{\tau \},*}}\left( Z_{\text{Mb}}^{\{t\}} |Z_{\text{s}}^{\{\tau \}}\right)$. The calculated $Z_{\text{a}}^{\{t\},*}$ is considered the optimal action for the robot to take in order to maximize progress or the information obtained.

\subsection{Simulation}\label{sec:Simulation}
To test the algorithm implemented for autonomous robot exploration, we performed simulations on the 35,126 2D floor plans available in the HouseExpo dataset utilising a modified version of the accompanying PseudoSLAM simulator \cite{li2019houseexpo}. The PseudoSLAM simulator is made to efficiently generate occupancy grid maps directly from 2D floor plans, without the computational burden of running a real SLAM algorithm. The simulator also calculates the percentage of the map that has been explored and keeps a count of the number of crashes. Thereby, the simulator is suitable for large-scale simulation studies. 

Unfortunately, the original PseudoSLAM simulator only allowed for the three fixed discrete movements: turn $\theta$ degrees to the left, turn $\theta$ degrees to the right, and move $X$ meters forward, where $\theta$ and $X$ are fixed variables. Thus, the original simulator was not suitable for the continuous movements calculated by \cref{eq:optimal_action}. Therefore, modifications were made to allow for such continuous movements in the simulator. Furthermore, it was found that the function "\textit{measure\_ratio()}" build into the PseudoSLAM simulator, meant to quantify the percentage of the map explored, counter-intuitively could return values greater than 1. Thus, we also modified this function. The modified PseudoSLAM simulator is available trough \cite{Damgaard_probMind_2022}. 

For our simulations, we adopted the simulation procedure used in \cite{li2019houseexpo}. One simulation with a random initial position was performed for each of the 35,126 2D floor plans. The simulations were limited to 200 time-steps. They were terminated if the "\textit{measure\_ratio()}" function returned more than $0.95$, corresponding to more than $95\%$ of the map had been explored. As an example, the result of one of the simulations is illustrated in \cref{fig:good_map_example}.

\begin{figure}[!ht]
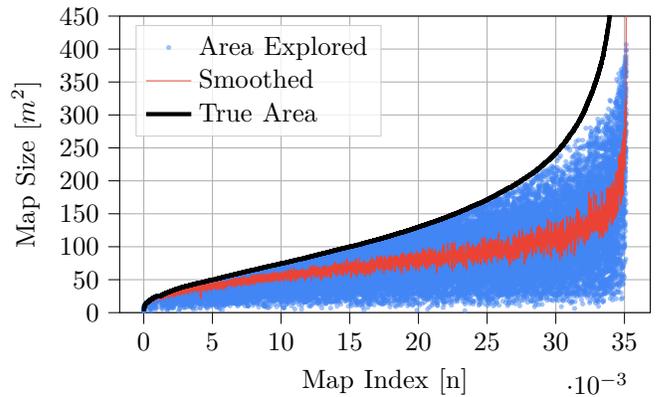

\centering
\includetikz{tikz/source_files/graphs}{map idx n - area explored m2}{0}
\caption{The area explored for each of the 35126 simulations performed with the indices sorted in ascending order by the true area of the map. The red curve shows a moving average with a windows size of 20.}\label{fig:mapID_vs_AREA}
\end{figure}

\begin{figure}[!ht]
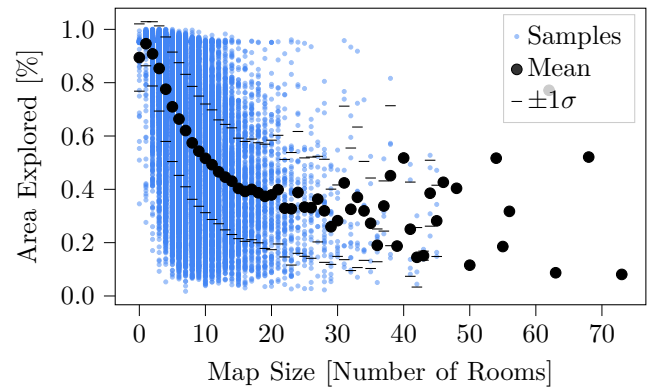

\centering
\includetikz{tikz/source_files/graphs}{map size N rooms - area explored procent}{0}
\caption{The percentage of area explored in each of the 35126 simulations performed compared to the number of rooms in each of the maps.}\label{fig:rooms_vs_area_explored}
\end{figure}

From \cref{fig:mapID_vs_AREA} it is seen that for the smallest floor plans in the data set, the robot manages to explore most of its environment. As the size of the floor plans increases, a smaller percentage of the environment is explored on average. This is expected behavior since there is a limit to how much of a map the robot can explore in a fixed amount of time steps. However, \cref{fig:rooms_vs_area_explored} might reveal another cause.

From \cref{fig:rooms_vs_area_explored} there seems to be a clear relationship between the number of rooms in the environment, and the percentage of the environment that the robot manages to explore. A possible cause of this could be that for the robot to explore multiple rooms it often has to pass through narrow doorways. Passing through narrow doorways presents a high risk of constraint violation. In many situations, there will be alternative paths away from doorways that still yield progress. Therefore, if the paths going through the doorway does not yield a high probability of information gain, the presented idiom will prefer actions away from such doorways. This means that the robot could spend more time-steps than necessary in rooms that are fully explored.

\begin{figure}[!ht]
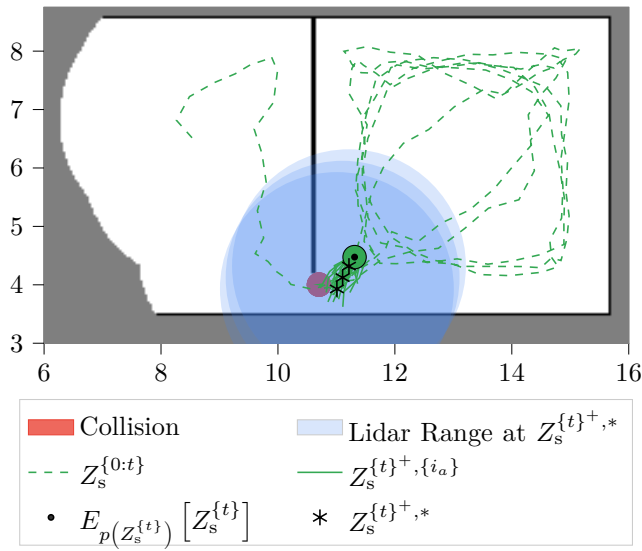

\centering
\includetikz{tikz/source_files/graphs/bad_map_example}{bad_map_example}{0}
\caption{The progress of a simulation for the map with ID "23e99dac3228ee2d371c5a627c49e415" after 200 time-steps. In this simulation, the robot spends a lot of time-steps driving around in the same room without getting out of it even though it is fully explored. The figure also shows an example of a collision in a doorway.}\label{fig:bad_map_example}
\end{figure}

As an example consider the simulation illustrated in \cref{fig:bad_map_example}. In this simulation, the robot starts in "room 1" and passes through a doorway to "room 2" already after a few time-steps. After passing through the doorway, the robot quickly explores the entire "room 2". However, since the area in "room 1" in close vicinity to the doorway is already explored, the probability of information gain for paths passing back through the doorway is low due to the limited lidar range used to define $p\left( Z_{\text{Pb}}^{\{\tau \}} |Z_{\text{s}}^{\{\tau \}} ,Z_{\text{LTM}}\right)$. Therefore, the robots keep driving around in "room 2" driven purely by progress. Overcoming this behavior would require some kind of memory about from which of the previous states the robot could obtain more knowledge, and some additional decision variables to guide the robot back to these states.

Besides guiding an agent towards new knowledge the idiom is also supposed to avoid constraints. In the implemented robot exploration algorithm, the only constraint is to prevent collisions with the robots surrounding. A total of 1617 unique collisions were recorded in 1253 different maps during the 6469065 time-steps simulated in all of the 35126 2D floor plans. Thus, only  $0,25\ \permille$ of the time-steps resulted in collisions. Nearly all of these collisions were registered near corners or doorways, like the collision shown in \cref{fig:bad_map_example}. Given that the idiom currently only supports checking constraints at discrete states, such behavior is to be expected, since the constraint can be satisfied at two consecutive states but not in between. Furthermore, for the specific application of robot exploration, this small probability of collision would probably be deemed tolerable, since in many cases would have to be a low-level collision avoidance system anyway. If this cannot be tolerated, the idiom would have to modified to include checking of constraint in between the discrete states.

Everything considered the ability of the idiom to guide an agent towards new knowledge while avoiding constraints seems to be as should be expected.

\section{Discussion}\label{sec:discussion}
In this paper, we have shown how to develop a generally applicable probabilistic programming idiom for the problem of making decisions under uncertainty to obtain new knowledge about an environment. We based our idiom on the memory structure of the Standard model of mind, and other ideas from research in cognitive architectures. We furthermore show how this idiom can be used for the specific problem of active mapping and robot exploration. Based on an extensive simulation study of this problem, it is concluded that the idiom works as could be expected. The simulation also indicated that the idiom probably would benefit from additional memory of old states in which more knowledge can be obtained. Furthermore, the simulation also indicated that the idiom for some application could benefit from checking constraints in between states.

\addtolength{\textheight}{-0cm}   

\bibliographystyle{IEEEtran}
\bibliography{IEEEabrv,bibliography}
\end{document}